\documentclass[10pt,twocolumn,letterpaper]{article}

\usepackage{cvpr}
\usepackage{times}
\usepackage{epsfig}
\usepackage{graphicx}
\usepackage{amsmath}
\usepackage{amssymb}
\usepackage[noend]{algpseudocode}
\usepackage{algorithmicx,algorithm}

\usepackage[breaklinks=true,bookmarks=false]{hyperref}

\cvprfinalcopy 

\def\cvprPaperID{4335} 

\begin{document}

\title{ NMS by Representative Region: Towards Crowded Pedestrian Detection by Proposal Pairing }

\author{Xin Huang$^1$\thanks{Both authors contributed equally to this work.} , Zheng Ge$^1{^*}$, Zequn Jie$^2$, Osamu Yoshie$^1$\\
$^1$Waseda University, $^2$Tencent AI Lab\\
{\tt\small koushin@toki.waseda.jp;jokerzz@fuji.waseda.jp;zequn.nus@gmail.com;yoshie@waseda.jp}}

\maketitle
\begin{abstract}
   Although significant progress has been made in pedestrian detection recently, pedestrian detection in crowded scenes is still challenging. The heavy occlusion between pedestrians imposes great challenges to the standard Non-Maximum Suppression (NMS). A relative low threshold of intersection over union (IoU) leads to missing highly overlapped pedestrians, while a higher one brings in plenty of false positives. To avoid such a dilemma, this paper proposes a novel Representative Region NMS (R$^2$NMS) approach leveraging the less occluded visible parts, effectively removing the redundant boxes without bringing in many false positives. To acquire the visible parts, a novel Paired-Box Model (PBM) is proposed to simultaneously predict the full and visible boxes of a pedestrian. The full and visible boxes constitute a pair serving as the sample unit of the model, thus guaranteeing a strong correspondence between the two boxes throughout the detection pipeline. Moreover, convenient feature integration of the two boxes is allowed for the better performance on both full and visible pedestrian detection tasks. Experiments on the challenging CrowdHuman~\cite{shao2018CrowdHuman} and CityPersons~\cite{zhang2017CityPersons} benchmarks sufficiently validate the effectiveness of the proposed approach on pedestrian detection in the crowded situation.
\end{abstract}

\begin{figure}[t]
\centering
\includegraphics[width=8cm]{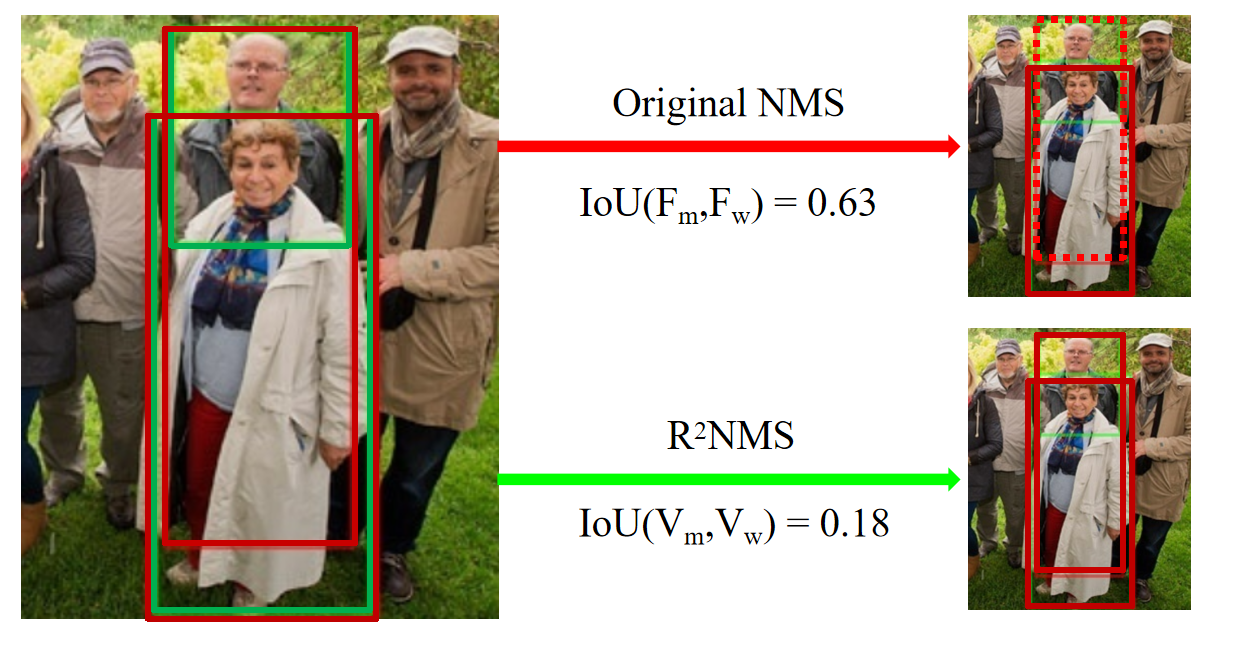}
\vspace{-0.2cm}
\caption{Illustration of R$^2$NMS. The image on the \textbf{left} shows two detected results before NMS. Red BBoxes are full body predictions and green BBoxes are visible body predictions. Two small images on the \textbf{right} show the final results which is processed by original NMS and R$^2$NMS. The red solid BBox represents the preserved BBoxes while red dotted BBox indicates the reduced true positive BBox. Arrows represent the IoU calculation. The IoU of their full body prediction is 0.63 while the IoU of their visible body is only 0.18. Thus, original NMS will reduce the red dotted BBox but R$^2$NMS is able to keep it.}
\vspace{-0.5cm}
\label{figure1} 
\end{figure}
\section{Introduction}

\label{sec:intro}

Pedestrian detection is a critical component of various real-world applications such as self-driving cars, and intelligent video surveillance. In recent years, the performance of pedestrian detectors has been rapidly improved with the rise of deep convolutional neural networks (CNNs)~\cite{simonyan2014very, he2016deep, hu2018squeeze}. However, pedestrian detection in the occluded situation remains challenging. Occlusion can be usually categorized into inter-class occlusion and intra-class occlusion. In inter-class occlusion, part of the human body is shielded by background objects such as pillar, car, trash box, and others. The feature of background objects confuses the model, leading to a high missing rate in this situation. A common solution to alleviate the inter-class occlusion is modeling based on instance parts~\cite{mathias2013handling, zhou2017multi, ouyang2013joint}. Visible parts can provide more discriminative and confident cues to guide the full-body detector. In intra-class occlusion, pedestrians have large overlaps with each other, so features of different instances will make detectors difficult to discriminate instance boundaries. As a result, detectors may give a lot of positives in overlapped area mistakenly. To solve this problem, Repulsion Loss~\cite{wang2018repulsion} and AggLoss ~\cite{zhang2018occlusion} propose additional penalties to the BBoxes which appear in the middle of the two persons. The proposals are forced to locate firmly and compactly to the ground-truth objects.

\begin{figure}[t]
\centering
\includegraphics[width=8cm]{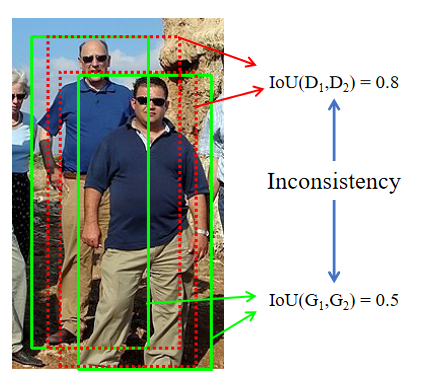}
\vspace{-0.2cm}
\caption{Illustration of the weakness of AdaptiveNMS~\cite{liu2019adaptive}. Green BBoxes are the full body annotations. Red dotted BBoxes are detected BBoxes which are not exactly match to ground truths. Overlap between the detected BBoxes is different from the overlap between ground truth BBoxes. If applying AdaptiveNMS in this situation. One detected BBox will be reduced. }
\vspace{-0.4cm}
\label{wadp} 
\end{figure}

However, even though the detectors succeed in identifying different human instances in a crowd, the highly overlapped results may also be suppressed by the post-processing of non-maximum suppression (NMS). This makes the current pedestrian detectors trapped in a dilemma: a lower threshold of intersection over union (IoU) resulting in the miss of highly overlapped pedestrians while a higher IoU threshold naturally brings in more false positives. To solve this problem, several modified versions of NMS have been proposed. Instead of directly discarding the highly overlapped BBoxes, soft-NMS~\cite{bodla2017soft} lowers detection scores of less confident BBoxes according to their overlaps with the most confident one. However, it still introduces lots of false positives of highly overlapped BBoxes. Adaptive NMS~\cite{liu2019adaptive} proposed a dynamic thresholding version of NMS. It predicts a density map, and sets adaptive IoU thresholds in NMS for different BBoxes according to the predicted density. However, density estimation itself remains a difficult task, and the exact matching from density to the optimal IoU threshold is also hard to decide. Moreover, the inaccurate BBox prediction often leads to the inconsistency between the ground-truth density and the IoU of the predicted BBoxes, as shown in Fig.~\ref{wadp}. This makes AdaptiveNMS still a sub-optimal solution.

In this paper, we propose a novel NMS algorithm to overcome the issues of existing NMS approaches called NMS by representative region (R$^2$NMS). R$^2$NMS leverages the visible parts of the pedestrians in NMS, which effectively averts the troubles brought by the difficult NMS on highly overlapped full bodies. Since the visible parts of pedestrians usually suffer much less from occlusion, a relative low IoU threshold sufficiently removes the redundant BBoxes locating the same pedestrian, and meanwhile avoids the large number of false positives. An illustration of R$^2$NMS is showing in Fig.~\ref{figure1}.

To obtain the visible part of a pedestrian, we propose a novel Paired-Box Model (PBM) based on the standard Faster R-CNN. PBM simultaneously predicts the full box and the visible box of a pedestrian in both RPN and the R-CNN module (\emph{i.e.}, from RoI sampling to the final post-classification and BBox regression layers). Specifically, a pair constituted by a full and a visible boxes is defined as the sample unit of both RPN and the R-CNN module. Such a pairing strategy guarantees a strong correspondence between the full and visible boxes throughout the detection pipeline. Moreover, the pairing solution allows the effective feature integration of the two boxes which benefits both the full and visible pedestrian detection tasks.

Experiments on the extremely crowded benchmark CrowdHuman~\cite{shao2018CrowdHuman} and the CityPersons~\cite{zhang2017CityPersons} show that the proposed approach can achieve the state-of-the-arts results, strongly validating the superiority of the method.

To summarize, the contributions of this work are three-fold: (1) a novel NMS method -- R$^2$NMS, to overcome the weakness of original NMS; (2) a Paired-Box Model (PBM) which simultaneously predicts both the full and visible boxes of a single pedestrian, and performs convenient feature integration of the two boxes; (3) the state-of-the-art results on the challenging CrowdHuman and CityPersons benchmarks.

\section{Related Works}
\noindent \textbf{Generic Object Detection.} With the rapid development of convolutional neural networks (CNNs)~\cite{simonyan2014very, he2016deep, hu2018squeeze}, great progress has been made in the object detection field. CNN based object detectors are usually categorized into one-stage and two-stage detectors. One-stage approaches~\cite{liu2016ssd, redmon2016you, lin2017focal, redmon2017yolo9000} aim to accelerate the inference process of detectors, to meet the requirement of time efficiency in various real world applications. In contrast to one-stage approaches, two stage detectors~\cite{girshick2015fast, ren2015faster, girshick2014rich} aim to pursue the cutting edge performance by adding a post classification and regression module to refine the detection results. To this end, Faster R-CNN~\cite{ren2015faster} together with its variants \emph{e.g.} FPN~\cite{lin2017feature} and Mask R-CNN~\cite{he2017mask} builds a powerful baseline for the task of generic object detection.

\begin{figure*}
\begin{center}
\includegraphics[width=18cm]{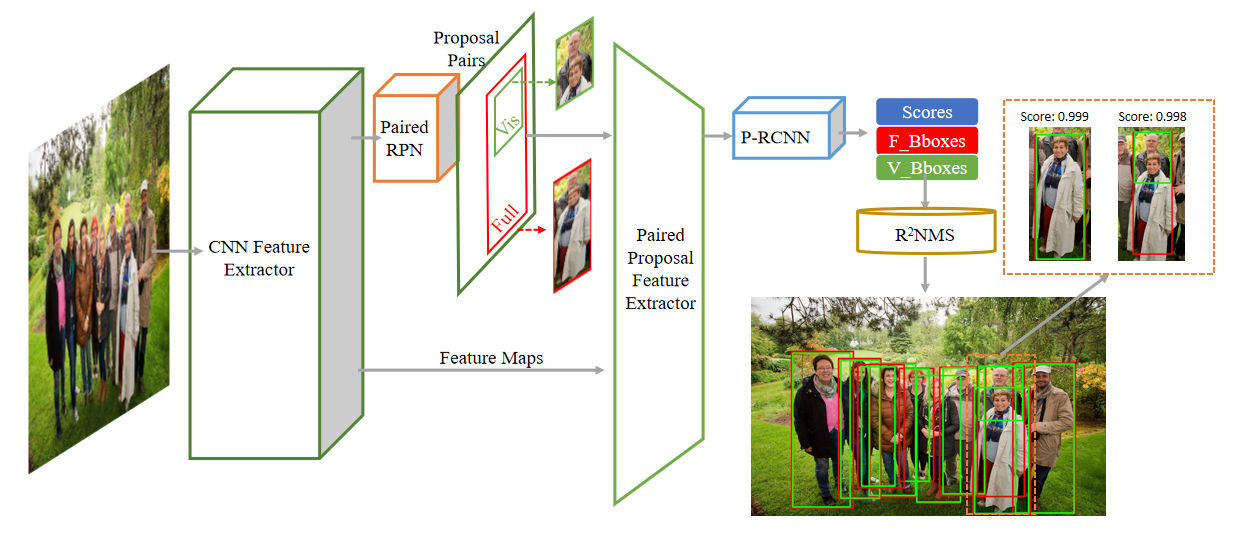}
\end{center}
\vspace{-0.4cm}
   \caption{Structure of our Paried-Box Model. Gray arrow represents the information flow. Pair RPN generates a pair of proposals from the same anchor. After that, pair proposal feature extractor aggregates the pair of proposal features and sends it to P-RCNN. P-RCNN predicts pairs of BBoxes. Finally, R$^2$NMS serves as our post-processing method to filter out false positives. Two paired BBoxes examples are shown in brown dotted box. }
\vspace{-0.4cm}
\label{model}
\end{figure*}

\noindent \textbf{Occlusion Handling for Pedestrian Detection.} Occlusion leads to two issues in pedestrian detection: mis-classifying occluded pedestrian and mis-placing detected results in a crowd. A common solution to the former problem is the part-based approaches~\cite{zhou2017multi, zhang2018occlusion, tian2015deep} which elaborate series of body-part detectors to handle the specific visual pattern of occluded instances. Except aforementioned methods, a few of recent works focus on utilizing annotations of visible body as extra supervisions to improve the performance of pedestrian detection. Zhou et al.~\cite{zhou2018bi} is the first one who regresses full and visible body of a pedestrian at the same time. Zhang et al.~\cite{zhang2018occluded} utilizes the annotations of visible parts as external guidance for better recognition performance on occluded instances.~\cite{pang2019mask} incorporates attention mechanism into pedestrian detection to force the detectors to focus more on visible regions of a pedestrian. Pedestrian detection in crowded scenes also raises a lot of attention.~\cite{wang2018repulsion} and ~\cite{zhang2018occlusion} both impose additional penalty terms on the BBoxes that appear in the middle of two persons. Adaptive NMS~\cite{liu2019adaptive} predicts a density map to perform a modified version of NMS with dynamic thresholds. ~\cite{zhang2019double} shares similar network architecture with our work, however, the regression performance for non-principle parts during inference stage can not be guaranteed, which makes Double Anchor RPN a sub-optimal solution. Different from all the existing works, our method utilizes the visible region information to assist the NMS, and forces the model to learn full and visible boxes together throughout the pipeline for the strong correspondence between the two boxes.

\section{Method}
\label{sec:Method}
In this section, we first analyze the weakness of the standard NMS and AdaptiveNMS~\cite{liu2019adaptive} when handling crowd situation. Next, we introduce the proposed R$^{2}$NMS. Finally, we describe the Paired-Box Model (PBM) in detail.
\vspace{-0.15cm}
\subsection{Analysis on original NMS and Adaptive NMS}
\vspace{-0.15cm}
In object detection, multiple object proposals locating the same object may be highly scored by the model. In this case, NMS is necessary to filter out the less confident ones according to the predicted scores. However, in the crowded situation, the ground-truth pedestrians are highly overlapped. Naturally, the detection boxes locating different pedestrians can also have high overlaps with each other. Therefore, when using a relative low IoU threshold as in the MS COCO benchmark~\cite{lin2014microsoft} during NMS, \emph{e.g.}, 0.5, many true positives of different instances may be suppressed. This significantly reduces the recall of all the instances, thus hurting the final detection performance. Take the CrowdHuman dataset as an example, for each of the 99,481 ground-truth instances in the validation set, we assume the detector can produce an exact BBox (\emph{i.e.}, the BBox is scored 1.0 and the IoU between the BBox and the ground-truth instance is also 1.0). However, after performing the standard NMS with IoU threshold 0.5, only 90,232 exact BBoxes are remained.\footnote{All the BBoxes are in a random order as their scores are all 1.0. Slightly different results are possible due to the random order in NMS.} Nearly $10\%$ of the ground-truth instances are missed in detection.  This indicates that even a perfect pedestrian detector fails to detect all the ground-truth instances, after the NMS using a relative low IoU threshold. On the contrary, setting a higher IoU threshold in NMS preserves more true positives, while significantly increase the false positives. Similarly, in the validation set of CrowdHuman, assuming all the ground-truth instances have exact predicted BBoxes, the missing rate will be reduced to $1\%$  when setting the IoU threshold of NMS as 0.7. However, the higher IoU threshold inevitably brings in more false positives in practice. For example, in the validation set of CrowdHuman, a well-trained Faster R-CNN based on ResNet-101 produces about 15,000 detection boxes whose score exceeds 0.5, after the NMS with the IoU threshold 0.7. Notice that the ground-truth instance number is 99,481, thus about 50,000 predicted boxes are redundant or false positives.  Therefore, the dilemma of the standard NMS in the crowded situation is difficult to resolve.

To overcome the shortcoming of the standard NMS, AdaptiveNMS~\cite{liu2019adaptive} is proposed. AdaptiveNMS~\cite{liu2019adaptive}  is a dynamic thresholding version of NMS. It incorporates a sub-network to predict the density for each location, and sets adaptive IoU thresholds in NMS for different BBoxes according to the predicted density. However, density estimation itself remains a difficult task. Besides, the matching from the density to the optimal IoU threshold is still hand-crafted in AdaptiveNMS, and thus the exact matching is difficult to acquire. Moreover, the inaccurate BBox prediction often leads to the inconsistency between the ground-truth density and the IoU of the predicted BBoxes. The phenomenon is illustrated in Fig.~\ref{wadp}. All these make the AdaptiveNMS still a sub-optimal solution.

\begin{figure}[t]
\centering
\includegraphics[width=8cm]{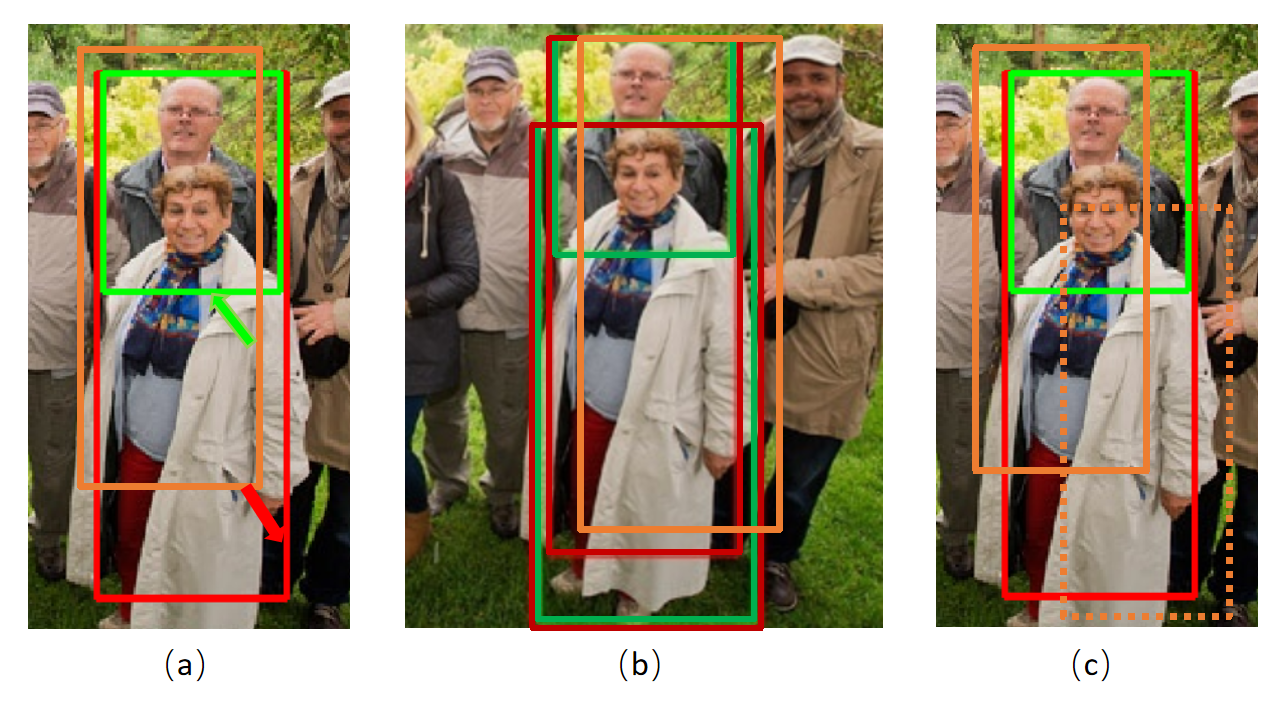}
\vspace{-0.25cm}
\caption{Illustration of P-RCNN. Red BBoxes represent the full body annotations. Green BBoxes stand for the visible annotations. Orange BBoxes are anchors. (a) P-RCNN regresses a pair of proposal from the same anchor. (b) The orange anchor is assigned to the man because it aligns well with the man's full and visible body. (c) Orange dotted anchor has large IoU with full body annotation, however, it is bad for visible body.}
\vspace{-0.6cm}
\label{P-RCNN} 
\end{figure}
\vspace{-0.15cm}
\subsection{NMS by Representative Region}
\vspace{-0.15cm}
To overcome the above issues of the standard NMS and AdaptiveNMS~\cite{liu2019adaptive} , we propose a novel NMS by representative region (R$^{2}$NMS). The key difference between R$^2$NMS and the standard NMS lies in the IoU calculation.  Specifically, instead of directly calculating the IoU of two full-body boxes as their ``overlap degree'', the IoU between the visible regions of the two boxes are used to determine whether the two full-body boxes are overlapped. Such a visible region based overlap determination is based on the following observations. BBoxes locating different pedestrians usually have low IoU between their visible regions, even if the IoU between the two full boxes is large. In contrast,  both the full and visible regions would have large IoUs when two BBoxes locate a same pedestrian. Therefore, the IoU between the visible regions of two boxes is a better indicator showing whether the two full-body boxes to belong to a same pedestrian. As a result, based on the visible regions, a relative low IoU threshold sufficiently removes the redundant BBoxes locating the same pedestrian, and meanwhile avoids the large number of false positives. The detailed algorithm of R$^2$NMS is described in Algorithm~\ref{alg:Algorithm1}.
\begin{algorithm}[t]
\caption{R$^2$NMS}
\label{alg:Algorithm1}
\hspace*{0.02in} {\bf Input:}\\
\hspace*{0.2in} Score : $S = \{s_1,s_2,...,s_n\}$\\
\hspace*{0.2in} Full body BBoxes : $B^{f} = \{b^{f}_1,b^{f}_2,...,b^{f}_n\}$\\
\hspace*{0.2in} Visible body BBoxes : $B^{v} = \{b^{v}_1,b^{v}_2,...,b^{v}_n\}$\\
\hspace*{0.2in} NMS threshold: $\Omega$\\
\hspace*{0.02in} {\bf Output:} \\
\hspace*{0.2in} Pairs of result : $R$
\begin{algorithmic}[1]
\State $R \leftarrow \{\}$\\
Removed BBoxes index list : $I \leftarrow \{\}$\\
According to $S$, rank $B^{f}$ and $B^{v}$ in descending order.
\For{$b^{v}_i \in B^{v}$}
　　\If{$i \in I\ or\ i=n$}
　　　　\State pass
　　\Else
      \State $j\leftarrow i+1$
　　　\For{$b^{v}_j \in B^{v}$}
          \State $overlap \leftarrow IoU(b^{v}_i,b^{v}_j)$
          \If {$overlap > \Omega$}
              \State Add $j$ to $I$
          \EndIf
      \EndFor
　　\EndIf
\EndFor
\For{$(b^{f}_i,b^{v}_i) \in (B^{f},B^{v})$}
    \If{$i \notin I$}
        \State Add $(b^{f}_i,b^{v}_i)$ to $R$
    \EndIf
\EndFor
\State \Return $R$
\end{algorithmic}
\end{algorithm}
\vspace{-0.15cm}
\subsection{Paired-BBox Faster R-CNN}
\vspace{-0.15cm}
To obtain  the  visible  part  of  a  pedestrian,  we  propose a  novel  Paired-Box  Model  (PBM)  which simultaneously predicts the full and visible boxes of a pedestrian. To this end, the PBM is based on a standard Faster R-CNN with the following  three modifications, \emph{i.e.}, Paired Region Proposal Network (P-RPN), Paired Proposal Feature Extractor (PPFE) and Pair R-CNN (P-RCNN). Specifically, the P-RPN first generates a set of full/visible proposal pairs, each of which corresponds to the full and visible regions of a pedestrian. PPFE then extracts the feature of each proposal pair, and fuses the features of the full and visible boxes to provide an integrated representation for each pair. Finally, the integrated representations are fed into P-RCNN to perform pair-wise classification and further refinement for the predicted full and visible BBoxes. In this manner, BBoxes of both full and visible body with strong correspondence can be obtained, facilitating the use of R$^{2}$NMS.

\noindent \textbf{Paired Region Proposal Network.}
The duty of Paired Region Proposal Network (P-RPN) is generating paired full-body and visible-body proposals. Since the full and visible regions of a pedestrian usually have high overlaps, it is feasible to regress a pair of full-body and visible-body proposals from a same anchor. Moreover, regressing the two proposals from a same anchor provides inherent correspondence between the predicted full-body and visible-body proposals.

The annotated full-body box $\rm F$ and the corresponding visible-body box $\rm V$ constitute a pair $\rm Q=(F,V)$, serving as the ground-truth unit of the model. We refer to the proposal matching method in ~\cite{zhou2018bi} to assign ground-truth label to the anchors during training P-RPN. More specifically,  the ground-truth assignment strategy in Faster R-CNN~\cite{ren2015faster} is modified by adding one more restriction. For a certain anchor, we consider both its IoU w.r.t. the full-body ground-truth boxes and its IoF w.r.t. the visible ground-truth boxes. Formally, an anchor $\rm A$ is viewed as positive matched to the ground-truth pair $\rm Q=(F,V)$, if the following requirements are satisfied.
$$\rm IoU(A,F)\ge\alpha_1\ and\ IoF(A,V)\ge\beta_1 $$
$$\rm IoU(A,F)= \frac{Area(A\cap F)}{Area(A\cup F)} $$
$$\rm IoF(A,V)= \frac{Area(A\cap V)}{Area(V)} $$
Here $\alpha_1$ and $\beta_1$ are positive thresholds for the full body and visible body, respectively. According to our experiments, PBM performs the best when $\alpha_1 = 0.7$ and $\beta_1 = 0.7$.

The detailed architecture of P-RPN follows the  RPN in Faster R-CNN~\cite{ren2015faster}. The only difference lies in the output layer. Apart from the locations of the paired proposals, P-RPN also predicts a score for each pair showing whether confidence of the pair matching to a pedestrian.  Therefore, for each dense anchor, P-RPN produces a 10-d  result ($\bf{R_{f}}$, $\bf{R_{v}}$, $\bf{S}$). Here $\bf{R_{f}}$  and $\bf{R_{v}}$ are  4-d BBox regression vectors ($f_{x}$,$f_{y}$,$f_{w}$,$f_{h}$) and ($v_{x}$,$v_{y}$,$v_{w}$,$v_{h}$), towards the full-body and visible-body ground-truths, respectively. $\bf{S}$ is a 2-d confidence vector ($S_{+}$, $S_{-}$) after softmax normalization. The loss functions used in training are the same as that in the standard RPN.

\begin{table*}[]
\vspace{-0.4cm}
    \caption{Main results. * stands for our re-implemented results. MR and AP are abbreviations of the Log-Average Missing Rate and the Average Precision, respectively. MR, AP, Recall stand for the results for full body. MR-V stands for the MR for visible body. For MR, lower is better. For AP and Recall, higher is better. $\Delta$ MR-V and  $\Delta$ MR-V indicate the absolute gain on visible and full body comparing to our re-implemented baseline. The best results are written in bold.}
\centering
\begin{tabular}{l|cccc|cccc|cc}
\hline
Method                      & P-RPN & P-RCNN & PPFE & R$^2$NMS  &MR-V          & MR             & AP         & Recall     &  $\Delta$ MR-V  &  $\Delta $MR \\ \hline
Baseline~\cite{shao2018CrowdHuman}     &- &- &- &-  & 55.94 & 50.42          & 84.95          & 90.24                                &          -       & -                             \\
Baseline*  & -&- &- &  -        & 55.57 & 46.28          & 84.91          & 88.25                                             &       -          &       -                      \\ \hline
NPM             &\multicolumn{1}{c}{$\surd$} &\multicolumn{1}{c}{$\surd$} &- &  -   & 54.18 &  45.43     & 85.59 & 88.92    &   +1.39       &   +0.85     \\
PBM             & \multicolumn{1}{c}{$\surd$}&\multicolumn{1}{c}{$\surd$} &mask & -& 52.70 & 44.20  & 85.60   & 88.61   &   +2.87        &   +2.08     \\
PBM             & \multicolumn{1}{c}{$\surd$}&\multicolumn{1}{c}{$\surd$} &mask &\multicolumn{1}{c|}{$\surd$} & 52.70 & \textbf{43.35}  & \textbf{89.29}   & \textbf{93.33}   &   +2.87        &   +2.93     \\
PBM             & \multicolumn{1}{c}{$\surd$}&\multicolumn{1}{c}{$\surd$} &concat &- &  \textbf{52.19} & 44.32  & 85.50   & 88.28   &   +3.38        &   +1.96     \\
PBM & \multicolumn{1}{c}{$\surd$} & \multicolumn{1}{c}{$\surd$} & concat & \multicolumn{1}{c|}{$\surd$} &  \textbf{52.19} & 43.57 & 89.28 & 93.10 & +3.38 & +2.71\\ \hline
\end{tabular}

\label{table1}
\end{table*}

\begin{figure}[t]
\centering
\includegraphics[width=8cm]{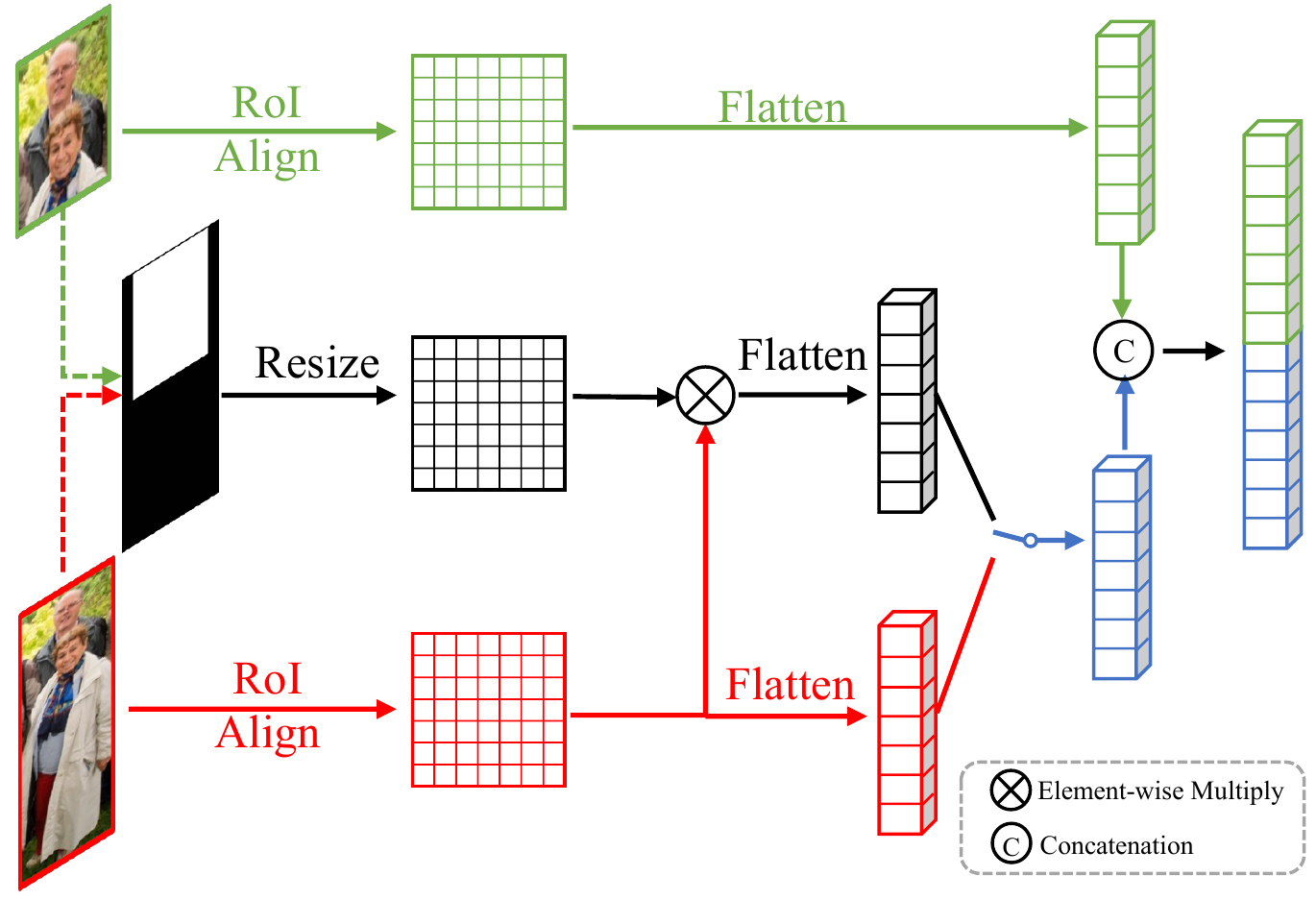}
\caption{Illustration of our proposed PPFE module.}
\vspace{-0.4cm}
\label{ppfe} 
\end{figure}

\noindent \textbf{Paired Proposal Feature Extractor.}
The pairing strategy in R-RPN allows an inherent correspondence between the full and visible proposals. We thus propose a Paired Proposal Feature Extractor (PPFE) to effectively integrate information from both proposals, facilitating the detection of both full and visible pedestrian detection tasks. 

 Figure~\ref{ppfe} shows two proposed ways of feature integration. A straightforward one is to directly concatenate the feature vectors of the full and visible proposals $\rm F_f$ and  $\rm F_v$ after RoI Align. Because only \emph{fc} layers are used in P-RCNN, spatial alignment of the features of the two proposals are not necessary. Experimental results show that such a simple feature fusion method increases the performance a lot.

The second way of feature integration is based on the attention mechanism which highlights the more informative visible regions. Specifically, for each pair of proposals, we generate a visible body attention mask showing whether to be inside the visible proposal for each pixel in the full-body proposal. If a pixel is also inside the visible proposal, we assign the value 1 to this pixel in the attention mask. Pixels outside the visible proposal are all set as 0 in the attention mask. The attention mask is then resized to the same size as the proposal feature after RoI Align, \emph{i.e.}, 7$\times$7.  We then  multiply the full-body proposal feature $\rm F_f$ with the mask in an element-wise manner, to get the visible mask attention feature $\rm F_m$. Finally, We concatenate $\rm F_v$ and $\rm F_m$ to obtain the final integrated feature for the pair.

\noindent \textbf{Paired R-CNN.}
The integrated feature produced by PPFE serves as the input of the Paired R-CNN (P-RCNN). P-RCNN is used to perform both full and visible pedestrian detection, based on the proposal pairs. The detailed architecture of P-RCNN is based on the R-CNN module in Faster R-CNN~\cite{ren2015faster}, with the following modifications. After receiving the pair feature, P-RCNN contains two bifurcated branches following two shared \emph{fc} layers for the full and visible BBoxes prediction, respectively. Each of the branches has the exact same architecture and output as that in the standard Faster R-CNN.

Similar to P-RPN, the essential problem in P-RCNN is how to assign the proposals to the ground-truths. We use a strategy which is quite similar to the anchor assignment method in P-RPN. To be more specific, for a pair of annotation $\rm Q=(F,V)$, a pair of proposal $\rm X=(P_f,P_v)$ is positive if it satisfies:
$$\rm IoU(P_f,F)\ge\alpha_2\ and\ IoU(P_v,V)\ge\beta_2 $$
According to our experimental results, the best number for $\alpha_2$, $\beta_2$ is 0.5 and 0.5.  The loss functions used in training are also the same as that in the standard Faster R-CNN.

As discussed above, the major modifications of PBM from Faster R-CNN introduce little extra computation, while brings a huge amount of performance gains. Experimental results in the next section verify the effectiveness of our model.
\begin{table}[]
\vspace{-0.4cm}
\caption{Influence of varying $\beta_1$ and $\beta_2$. NPM is the abbreviation of Naive Pair Model. The value of MR-V+MR reflects the model's performance on two of the annotation categories.}
\centering
\begin{tabular}{ccc|ccc}
\hline
Method &$\beta_1$ &$\beta_2$ & MR-V & MR & MR-V+MR  \\ \hline
Baseline  & - &  -    & 55.57 & 46.28  &  101.85   \\ \hline
NPM   &0.8 &0.5 & 54.81 & 46.34     &     101.15         \\
NPM   &0.6 &0.5 & 54.65 & 45.42      &    100.07   \\
NPM   &0.7 &0.5 & \textbf{54.18} & 45.30 &  \textbf{99.48}  \\
NPM   &0.7 &0.6 & 54.35 & 47.23     &     101.58   \\
NPM   &0.7 &0.4 & 55.80 & \textbf{44.53} &  100.33  \\ \hline
\end{tabular}
\vspace{-0.4cm}
\label{table2}
\end{table}
\section{Experiments}
To evaluate our proposed methods, we conduct several experiments on two crowd pedestrian datasets: CrowdHuman \cite{shao2018CrowdHuman} and CityPersons \cite{zhang2017CityPersons}.
\subsection{Datasets and Evaluation Metric}
\noindent \textbf{CrowdHuman Dataset.}
Recently, CrowdHuman \cite{shao2018CrowdHuman} dataset, a human detection benchmark, has been released to better evaluate pedestrian detectors in crowded scenarios. There are 15000, 4370 and 5000 images in training set, validation set, and test set respectively. The average of the number of persons in an image is 22.6. CrowdHuman \cite{shao2018CrowdHuman} provides three categories of bounding boxes annotations for each human instance: head bounding-box, human visible-region bounding-box and human full-body bounding-box. All of our experiments are conducted under the settings of full body and visible body. The models are trained on the training set and evaluated on validation set.

\noindent \textbf{CityPersons Dataset.}
The CityPersons \cite{zhang2017CityPersons} dataset is a subset of Cityscapes \cite{cordts2016cityscapes} which only consists of person annotations. There are 2975 images for training, 500 and 1575 images for validation and testing. The average of the number of pedestrians in an image is 7. The visible-region and full-body annotations are provided. We evaluate our proposed methods under the full-body setting. Following to the evaluation protocol in CityPersons \cite{zhang2017CityPersons}, objects whose height are less than 50 pixels are ignored. The validation set is further divided into several subsets according to visibility:

(1) Reasonable (\textbf{R}): $\rm Visibility \in [0.65, \infty)$

(2) Heavy Occlusion (\textbf{HO}): $\rm Visibility \in [0.2, 0.65)$

We show our results across these two subsets.

\noindent \textbf{Evaluation Metric.}
For evaluation, we follow the standard Caltech \cite{dollar2011pedestrian} evaluation metric -- MR, which stands for the Log-Average Missing Rate over false positives per image (FPPI) ranging in $[10^{-2}, 10^{0}]$. To better evaluate our methods, Average Precision (AP) and Recall are also provided.

\begin{table}[]
\vspace{-0.4cm}
\caption{Ablation study on the PPFE module.}
\centering
\begin{tabular}{c|c|cccc}
\hline
Method    & PPFE & MR-V           & MR             & AP             & Recall         \\ \hline
NPM     & -  & 54.18          & 45.43          & 85.59          & 88.92 \\
PBM & concat &  \textbf{52.19}          & 44.32         & 85.50          & 88.28          \\
PBM  & mask &  52.70              &  \textbf{44.20} & 85.50          & 88.61          \\ \hline
\end{tabular}
\vspace{-0.4cm}
\label{table3}
\end{table}

\subsection{Implementation Details}
For CrowdHuman \cite{shao2018CrowdHuman}  dataset, we adopt Feature Pyramid Network (FPN) \cite{lin2017feature} with ResNet-50 \cite{he2016deep} as our baseline. To extract more precise features, we adopt RoI Align \cite{he2017mask} instead of RoI Pooling \cite{ren2015faster} for feature extraction. The anchor aspect ratios for full body and visible body are set to [0.5,1,2]. Because images in CrowdHuman dataset have various shapes, we resize them so that the short edge is 800 pixels while the long edge is smaller than 1400 pixels. We train our model on 8 GPUs with totally 16 images per mini-batch. We use SGD with momentum of 0.9 as our optimizer and set the initial learning rate as 0.02. We train 20 epochs in total and decrease the learning rate by 0.1 at 16th and 19th epochs.

For CityPersons dataset, we follow the settings in adapted Faster R-CNN framework \cite{zhang2017CityPersons}. Specifically, the backbone of our detector is VGG-16 \cite{simonyan2014very}. To detect small objects, we remove the fourth max-pooling layer in VGG-16. The aspect ratio for anchor is set to 2.44. The anchor sizes are the same as in \cite{zhang2017CityPersons}. We also adopt Adam as our optimizer. We train our model 12 epochs in total on 8 GPUs with a total of 16 images per mini-batch. The initial learning rate is set to 0.0008. We decrease the learning rate by 0.1 at the 8th and 11th epochs. We do not upsample the input images and only use the reasonable subset of pedestrians for training. 


\subsection{Detection Results on CrowdHuman}

\noindent \textbf{Main results.} To thoroughly evaluate the performance of our proposed methods, we conduct plenty of experiments on CrowdHuman~\cite{shao2018CrowdHuman} dataset and evaluate the performance under three evaluation metrics. MR is chosen as the main metric. Table \ref{table1} shows the performance of baseline and our proposed methods on CrowdHuman~\cite{shao2018CrowdHuman} validation subsets. For fair comparison, all the models listed in Table \ref{table1} share the same settings on hyper-parameters. As can be seen in Table \ref{table1}, our re-implemented FPN~\cite{lin2017feature} baseline achieves 46.28\% MR on full body detection and 55.57\% MR on visible body detection, which outperform the baseline in CrowdHuman~\cite{shao2018CrowdHuman} by 4.14\% and 0.37\%, respectively. Therefore, our baseline is strong enough to validate the effectiveness of our proposed methods. Based on our strong baseline, our methods can further bring the notable 2.71\%, 4.37\% and 4.85\% improvements on MR, AP and Recall, respectively, which significantly demonstrate the ability of our methods.  To analyze the contribution of our proposed modules individually, we progressively replace the components in the baseline model with our modules. The relevant ablation studies and analysis are illustrated in the following paragraphs.

\begin{table}[]
\vspace{-0.4cm}
\caption{Ablation study on R$^2$NMS. R$^2$ is the abbreviation of R$^2$NMS. PBM here utilizes the concat version of PPFE.}
\begin{tabular}{l|cc|cccc}
\hline
Method         &$\beta_1$ &$\beta_2$ & MR-V  & MR    & AP    & Recall \\ \hline
NPM            & 0.7 & 0.4 & 55.80  & 44.53 & 85.62 & 88.74   \\
NPM+R$^2$      & 0.7 & 0.4 & -     & 45.91 & 88.21 & 93.27  \\\hline
NPM            & 0.7 & 0.5 & 54.18 & 45.30& 85.59 & 88.69           \\
NPM+R$^2$      & 0.7 & 0.5 & -     & 45.87 & 88.39 & 93.76  \\\hline
PBM            & 0.7 & 0.5 & 51.92 & 44.79 & 85.62 & 88.12  \\
PBM+R$^2$      & 0.7 & 0.5 & -     &43.57 & 89.28 & 93.10  \\ \hline
\end{tabular}
\vspace{-0.4cm}
\label{table4}
\end{table}
\noindent \textbf{Influence of different hyper-parameter settings in pair-based sample unit.}
To evaluate the efficacy of our pair-based sample unit in the whole pipeline, we build a naive version of paired model which simply replace the RPN and R-CNN in Faster R-CNN by P-RPN and P-RCNN. In NPM, full and visible proposals from RPN are fed into two separate R-CNN modules. These two separate R-CNN modules are responsible for the full and visible body detection, respectively. Each R-CNN module has the same architecture as the standard one in Faster R-CNN. We conduct a group of experiments on NPM to find out the most reasonable hyper-parameter settings. To make the NPM have consistent hyper-parameter setting with the baseline, $\alpha_1$ and $\alpha_2$ are fixed, \emph{i.e.}, 0.7 and 0.5, separately. The influence of the varying $\beta_1$ and $\beta_2$ is comprehensively studied and the results are shown in Table \ref{table2}. We can see that NPM works reasonably well with $\beta_1=0.7$ and $\beta_2=0.5$. When $\alpha_1$ and $\alpha_2$ are fixed, $\beta_1$ and $\beta_2$ control the balance between visible body and full body in our pair-based sample unit for training. When $\beta_1$ and $\beta_2$ are small, more low quality visible proposals are counted, which hurts the performance of NPM on visible detection results. A large $\beta_1$ and $\beta_2$ exclude poorly aligned visible proposals, meanwhile, such a setting rejects some well aligned full body positive training examples, which leads to a poor full body detection result. As can be seen in the third row of Table \ref{table2}, $\beta_1=0.7$ and $\beta_2=0.5$ achieve a good trade-off between the two annotation categories. Therefore, we use $\alpha_1=0.7$, $\alpha_2=0.5$, $\beta_1=0.7$ and $\beta_2=0.5$ in the whole experiments unless otherwise mentioned. 

\begin{table}[t!]
\vspace{-0.4cm}
\caption{State of the art comparison on CrowdHuman. * stands for our re-implemented results. The PBM here is equipped with mask version of PPFE.}
\centering
\begin{tabular}{l|ccc}
\hline
Method       & MR    & AP    & Recall   \\ \hline
Baseline (CrowdHuman)        & 50.42          & 84.95          & 90.24    \\
Baseline*          & 46.28          & 84.91          & 88.25        \\
AdaptiveNMS         & 49.73 & 84.71 & 91.27    \\
Repulsion Loss*         & 45.69 & 85.64 & 88.42    \\
PBM  & \textbf{43.35} & \textbf{89.29} & \textbf{93.33}    \\ \hline
\end{tabular}
\label{table5}
\vspace{-0.4cm}
\end{table}
\noindent \textbf{Impact of PPFE.} When equipping NPM with the PPFE module, it becomes the PBM. We compare PBM with NPM in Table \ref{table3}. From Table \ref{table3}, we can tell that both proposed feature integration methods in Section \ref{sec:Method} bring remarkable improvements on MR-V and MR. The simpler one -- direct concatenating the features of full and visible proposals improves MR-V and MR by 1.99\% and 1.11\%, respectively, while PPFE with attention mechanism shows even better performance. Such a large gap between PBM and NPM totally proves that our proposed PPFE module could extract and integrate features from pair-based sample unit efficiently and successfully. Moreover, from Table \ref{table4}, we can learn that PPFE makes R$^2$NMS to perform better by improving the model's performance on visible body, illustrating the necessity of PPFE module. 


\noindent \textbf{The relation between MR-V and R$^2$NMS.} To demonstrate the effectiveness of R$^2$NMS, we replace the original NMS with the proposed R$^2$NMS. Table \ref{table4} shows three groups of experimental results. We can conclude that R$^2$NMS can boost performance on AP and Recall under all settings, while R$^2$NMS makes MR worse when NPM is applied. To understand why R$^2$NMS weakens the MR of the NPM, we would like to introduce the relation between MR-V and R$^2$NMS. It is natural to believe that the quality of visible body predictions is crucial because R$^2$NMS uses the IoU between the visible regions of two BBoxes to determine whether two full-body BBoxes are overlapped. However, we argue that compared to \emph{absolute localization quality} of predicted visible BBoxes, \emph{relative localization quality} which captures overlap degree between two human instances is more important. A straightforward example can verify our point -- suppose all visible body predictions are the exactly same as full body predictions, in this case, MR-V will be poor. However, using such visible body predictions during R$^2$NMS leads to the exactly same results for full body detection as original NMS, which is not as poor as MR-V. This example explicitly shows that a poor MR-V does not necessarily lead to poor MR for full body predictions during R$^2$NMS. 

To be more concrete, although a lower MR-V can possibly better model the overlaps between each instance pair, which will further benefit full body detection, as stated above, the absolute value of MR-V is not the decisive factor to the performance of full body detection. Our experimental results in Table \ref{table4} also demonstrate this point. With the decrease of MR-V, the performance of R$^2$NMS gets better and a MR-V under 54\% is good enough to bring positive effects on full body detection via R$^2$NMS. 

\begin{table}[t]
\vspace{-0.4cm}
\caption{Comparison between our PBM and baseline on the CityPersons~\cite{zhang2017CityPersons}. Log-Average Missing Rates (MR) on validation subsets is reported. \textbf{R} refers to reasonable set while \textbf{HO} refers to heavy occlusion set.}
\begin{tabular}{c|cc|c|c}
  \hline
Method   & \multicolumn{1}{c}{PPFE} & \multicolumn{1}{c}{R2NMS} & \multicolumn{1}{|c}{\textbf{R}} & \multicolumn{1}{|c}{\textbf{HO}} \\\hline
Baseline & -                        &            -               & 13.8                           & 59.0                            \\\hline
PBM  & concat                   &             -              & 12.5                           & 57.3                            \\
PBM  & concat                   & \multicolumn{1}{c|}{$\surd$} & 12.1                           & 57.0                         \\
PBM  & mask                     &              -             & 12.3                           & 54.9                            \\
PBM  & mask                     & \multicolumn{1}{c|}{$\surd$} & \textbf{11.1}                  & \textbf{53.3}                \\\hline
\end{tabular}
\vspace{-0.4cm}
\label{table6}
\end{table}
\noindent \textbf{Further Analysis on R$^2$NMS.} The experimental results in Table \ref{table1} show that R$^2$NMS can significantly improve AP and Recall, while it only improves MR by less than 1 percent. Such a phenomenon is caused by the difference between MR and AP, which we would like to deeply discuss. 

The main difference between MR and AP lies in the range of interested predicted scores. MR only cares about the predicted BBoxes whose scores are higher than the highest scored false positive. In contrast, AP takes all the detection results scored between 0 to 1 into consideration. Therefore, only a small fraction of predicted results will affect MR. As discussed in Section \ref{sec:intro} and Section \ref{sec:Method} , in crowded situation, detectors tend to generate a lot of highly scored false positives which are difficult to remove via NMS. Thus, we claim that a large number of highly scored false positives in the results make FPPI reach one quickly. To demonstrate this, we calculate the average score of the highest scored false positive across all testing images. It turns out that such value is extremely high, and sometimes is even beyond 0.9. Such a phenomenon indicates that only BBoxes whose scores are higher than 0.9 can affect MR. Therefore, MR reflects the performance of the highly scored fraction of the detection results, while AP measures the performance of all detected BBoxes. This difference leads to the huge discrepancy between the gain on MR and AP from R$^2$NMS.


In conclusion, although R$^2$NMS does not bring large improvements on MR, it greatly boosts AP. The huge improvement on AP strongly validates that R$^2$NMS can not only preserve more true positives comparing to the original NMS, but also introduce less false positives.

\begin{table}[!t]
\vspace{-0.4cm}
\caption{State of the art comparison on CityPersons. AdaptiveNMS+ refers to AdaptiveNMS with AggLoss. * represents our implementation.}
\begin{tabular}{l|c|c|c}
\hline
Method             & \multicolumn{1}{c}{Backbone} & \multicolumn{1}{|c}{\textbf{R}}& \multicolumn{1}{|c}{\textbf{HO}} \\ \hline
Baseline*          & VGG-16                       & 13.8       & 59.0                            \\
Baseline (MGAN)\cite{pang2019mask}     & VGG-16                       & 13.8       & 57.0                            \\\hline
Adapted FasterRCNN\cite{zhang2017CityPersons} & VGG-16                       & 15.8       & -                               \\
ATT-part\cite{zhang2018occluded}           & VGG-16                       & 16.0       & 56.7                            \\
Repulsion Loss\cite{wang2018repulsion}     & ResNet-50                    & 13.2       & 56.9                            \\
OR-CNN\cite{zhang2018occlusion}             & VGG-16                       & 12.8       & 55.7                            \\
AdaptiveNMS\cite{liu2019adaptive}        & VGG-16                       & 12.9       & 56.4                            \\
AdaptiveNMS+\cite{liu2019adaptive}       & VGG-16                       & 11.9       & 55.2                            \\
MGAN\cite{pang2019mask}               & VGG-16                       & 11.5       & 51.7                            \\\hline
Ours               & VGG-16                       & \textbf{11.1}       & 53.3                            \\\hline
\end{tabular}
\vspace{-0.4cm}
\label{table7}
\end{table}
\vspace{-0.15cm}
\subsection{State-of-the-art Comparison on CrowdHuman}
\vspace{-0.15cm}
We compare our method with AdaptiveNMS \cite{liu2019adaptive} and Repulsion Loss \cite{wang2018repulsion} on the CrowdHuman validation set in Table \ref{table5}. It clearly shows that PBM with R$^2$NMS outperforms these two published methods. Our method significantly reduces MR from 49.73\% to 43.35\% and boosts AP from 84.71\% to 89.29\%. Such a large gap demonstrates the superiority of our PBM and R$^2$NMS.

\subsection{Detection Results on CityPersons}
To prove the generalization ability of our methods, we also conduct several experiments on CityPersons \cite{zhang2017CityPersons}. Comparisons results are shown in Table \ref{table6}. To clearly demonstrate the effectiveness of our proposed new components, we also show the performance of out method with different settings in different rows. No matter under what kind of setting, PBM with R$^2$NMS performs better than baseline. The best results are boldfaced and shown in the final row. Compared to the baseline, our detector significantly reduces the MR from 13.8\% to 11.1\% on the reasonable set. On the heavy occlusion set, it outperforms the baseline by 5.7\%. Such a large gain provides a convincing proof to the effectiveness of our detector.

We compare our method with the recent state-of-the-art methods including Adapted FasterRCNN \cite{zhang2017CityPersons}, ATT-part \cite{zhang2018occluded}, Repulsion Loss \cite{wang2018repulsion}, OR-CNN \cite{zhang2018occlusion}, AdaptiveNMS \cite{liu2019adaptive} and MGAN \cite{pang2019mask} on the CityPersons validation set. We list the performance of previous works on reasonable subsets with the original input size in Table \ref{table7}. We evaluate our methods under the same settings. The proposed PBM with R$^2$NMS outperforms all published methods on reasonable validation subsets. Our method reduces the MR of state-of-the-art result from 11.5\% to 11.1\%. Notice that our method performs slightly worse than MGAN due to the weaker baseline result of our model.

\section{Conclusion}
In this paper, we propose R$^2$NMS to effectively remove the redundant boxes without brings in many false positives in crowded situation. The R$^2$NMS uses the IoU between the visible regions of the two boxes to determine whether the two full-body boxes are overlapped. To support this idea, we propose a novel Paired-Box Model (PBM) to simultaneously predicts the full box and the visible box of a pedestrian. Experiments on the extremely crowded benchmark CrowdHuman~\cite{shao2018CrowdHuman} and the CityPersons~\cite{shao2018CrowdHuman} show that the proposed approach can achieve the state-of-the-arts results, strongly validating the superiority of the method.

{\small
\bibliographystyle{ieee_fullname}
\bibliography{egbib}
}

\end{document}